\pdfoutput=1

\documentclass[11pt]{article}

\usepackage[]{EMNLP2023}

\usepackage{times}
\usepackage{latexsym}

\usepackage[T1]{fontenc}

\usepackage[utf8]{inputenc}

\usepackage{microtype}

\usepackage{inconsolata}

\usepackage{graphicx,multirow}
\usepackage{amsmath}
\usepackage{comment}

\usepackage{algorithm}
\usepackage{algpseudocode}

\title{Targeted Image Data Augmentation \\ Increases Basic Skills Captioning Robustness}

\newcommand*{\affaddr}[1]{#1} 
\newcommand*{\affmark}[1][*]{\textsuperscript{#1}}
\newcommand*{\email}[1]{\texttt{#1}}

\author{%
Valentin Barriere\affmark[1,2]*, Felipe del Rio\affmark[1,3]*, Andres Carvallo de Ferari\affmark[2]*, \\
\textbf{Carlos Aspillaga},\affmark[1] \textbf{Eugenio Herrera-Berg},\affmark[1] \textbf{Cristian Buc}\affmark[1]\\
\affaddr{\affmark[1]Centro Nacional de Inteligencia Artificial, Macul, Chile}\\
\affaddr{\affmark[2]Department of Computer Science, Universidad de Chile, Santiago, Chile}\\
\affaddr{\affmark[3]Department of Computer Science, Pontificia Universidad Catolica, Santiago, Chile}\\
\email{name.lastname@cenia.cl}\\
}

\begin{document}
\maketitle
\begin{abstract}
Artificial neural networks typically struggle in generalizing to out-of-context examples. One reason for this limitation is caused by having datasets that incorporate only partial information regarding the potential correlational structure of the world. In this work, we propose TIDA (Targeted Image-editing Data Augmentation), a targeted data augmentation method focused on improving models' human-like abilities (e.g., gender recognition) by filling the correlational structure gap using a text-to-image generative model. More specifically, TIDA identifies specific skills in captions describing images (e.g., the presence of a specific gender in the image), changes the caption (e.g., "woman" to "man"), and then uses a text-to-image model to edit the image in order to match the novel caption (e.g., uniquely changing a woman to a man while maintaining the context identical). Based on the Flickr30K benchmark, we show that, compared with the original data set, a TIDA-enhanced dataset related to gender, color, and counting abilities induces better performance in several image captioning metrics. 
Furthermore, on top of relying on the classical BLEU metric, we conduct a fine-grained analysis of the improvements of our models against the baseline in different ways. 
We compared text-to-image generative models and found different behaviors of the image captioning models in terms of encoding visual encoding and textual decoding.\footnote{Code will be available online after submission.}

\end{abstract}

\begin{figure*}[!h]
    \centering
    \includegraphics[width=.8\textwidth]{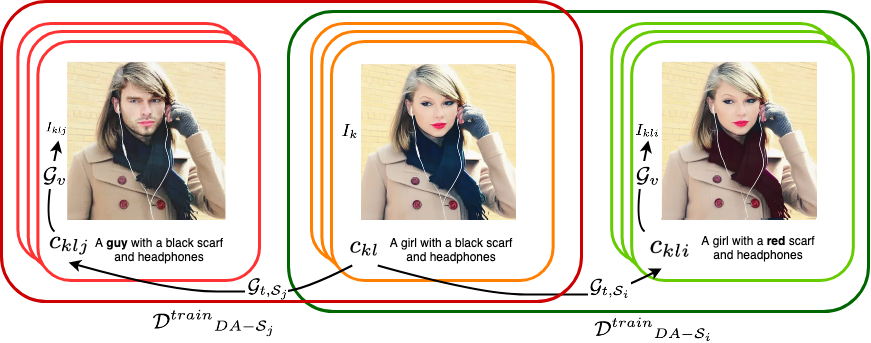}
    \caption{TIDA Framework (Example generated with Null-Text-Inversion \cite{Mokady2022})} 
    \label{fig:tida}
\end{figure*}

\section{Introduction}

Humans and animals develop all kinds of cognitive abilities from a very early age that allow them to interact with their world \cite{spelke1992origins,spelke2007core}. For instance, infants display numerical cognition abilities \cite{feigenson2004core,xu2000large}, can recognize emotions \cite{bornstein2003recognition} or even the danger associated with other agents' action plans \cite{liu2022dangerous}. Comparatively, animals also display similar numerical cognition abilities \cite{davis1982counting,dacke2008evidence}, or recognize emotions in order to better communicate within a social group \cite{hantke2018my}. These abilities are crucial in order to build models of the world that are necessary for planning, reasoning, and solving complex decision-making tasks \cite{lake2017building}. 

Deep learning systems can solve these tasks by optimizing an objective function via supervised, semi-supervised or unsupervised learning \cite{lecun2015deep}. Within this framework, it has been shown that deeper layers progressively represent increasingly abstract concepts \cite{krizhevsky2017imagenet}, akin to what has been observed in the human visual or auditive processing pathways \cite{cichy2016comparison,caucheteux2023hierarchical}. 
Moreover, empirical work has shown that pretrained state-of-the-art transformer models \cite{devlin2018bert} encode factual knowledge within sets of knowledge neurons \cite{Dai2022}; strongly related to the concepts of "grandmother" cells in neuroscience \cite{quiroga2005invariant}. Importantly, not only factual knowledge but also conceptual knowledge (such as "sentiment" in a text or "written language" in an image) are encoded by nodes in deep layers \cite{Radford2017,yosinski2015understanding}. Whereas recent methods have been proposed to access and edit factual knowledge \cite{Meng2022}, and thus evaluate how and where facts are being encoded in deep networks \cite{Meng2022a}, it is much harder to evaluate the abilities associated with conceptual knowledge stored in these networks. Yet, possessing such a conceptual knowledge base is crucial for out-of-distribution generalization \cite{bosselut2019comet}.

Although deep networks seem to encode conceptual knowledge that allows them to display human-like abilities such as counting, emotion, gender, color, and sentiment recognition/categorization \cite{Wallace2019numbers,Barriere2022opinions,Hendricks2018,Anderson2016,Barriere2017b}, these same networks typically struggle in producing out-of-context (or out-of-distribution) generalizations \cite{marcus2018deep,lake2018generalization,ruis2020benchmark,delrio2023studying,ribeiro2020beyond}. 
%
These limitations are due to the inherent functioning of Artificial Neural Networks (ANNs). Indeed, generalization performances of ANNs largely depend on their ability to extract the correlational structure in the training data set, memorize this structure, and extrapolate it to a novel (test) data set \cite{krizhevsky2017imagenet,saxe2019mathematical}. 
Indeed, given that the performance of vanilla deep networks is constrained by the structural correlation observed in the training data set, a straightforward way to maximize the generalization performance in ANNs is to augment data sets in \textit{targeted} ways \cite{sharmanska2020contrastive,He2023}. 
Thereby, targeted data augmentation would increase the span of potential correlations that could be observed in the world, and as such improve the human-like abilities of deep networks. By targeting specific human-like abilities and augmenting the data set to encapsulate unseen examples associated with these abilities, we hypothesize that models can increase their conceptual knowledge, and thus improve their performance on specific benchmarks we discuss below. 
Moreover, similar to editing unique factual knowledge \cite{Meng2022}, one would ideally want to target unique conceptual knowledge (e.g., gender, color, numerosity, emotion, shape...) to induce such ability-selective performance, which has been widely studied \cite{anderson2016spice,Hu2023}. 

We will propose a simple way to overcome the issues raised above, for Image Captioning (IC) task. 
Interestingly, novel text-to-image generation models \cite{rombach2022high} in combination with text-generation or manipulation \cite{He2023,mitkov2022oxford,Murty2022} 
affords novel possibilities for targeted data augmentation for vision-language tasks. 
%
Hence, we propose to enhance the capabilities of an Image Captioning model by using a targeted data-augmentation on several specific abilities (or skills). We use simple regular expressions (regex) to identify these skills in the caption, to change the caption for another version of it, and to generate the image related to this caption.  
%
The main contributions of this work are twofold. First, we propose a simple method to identify data related to a specific human-like ability in image captioning (e.g., color identification, emotion recognition...). Second, we propose a novel data augmentation method based on image-to-text generation models that allows one to generate data sets that can selectively improve a single or combinations of human-like skills in image captioning performance. Instead of manipulating or fine-tuning information processing within image captioning models, our method increases the span of potential object correlations and thus allows us to generalize image captioning abilities to a broader spectrum of situations that can be observed in the real world \cite{zhang21}. In what follows, we first describe related work while specifying the original contribution of our work. Subsequently, we describe the Targeted Image-editing Data Augmentation (TIDA; see Figure \ref{fig:tida}) method and present the results associated with fine-tuning models with our TIDA-augmented data sets. Finally, we discuss the implications of our work.

\section{Previous and Related Work}



%




\paragraph*{Image Captioning}
Image captioning (IC) models provide human-like captions to images \cite{cornia2020meshed,herdade2019image}. Such an ability lies in the intersection between computer vision and natural language processing \cite{devlin2015language}, and is therefore, in essence, a multimodal problem. Early IC models proposed to sequentially combine convolutional neural networks (CNN) with recurrent neural networks (RNN) into a single imaged-conditioned language model \cite{karpathy2015deep,chen2015mind,fang2015captions}. Given the success of these models and their potential industrial applications, subsequent work has focused on improving the models' image captioning ability by focusing on specific properties of IC models. For instance, it has been shown that top-down visual attention mechanisms improve captioning performance \cite{anderson2018bottom,lu2017knowing}. Alternatively, focusing on the learning process, it has been shown that implementing self-critical sequence training (a variant of the REINFORCE algorithm) improves IC performances by avoiding the exposure bias \cite{ranzato2015sequence} and directly optimizing the relevant task metrics \cite{rennie2017self}. Furthermore, many IC models are pre-trained using tasks like Masked Language Modeling (MLM) and Image-Text Matching (ITM). These tasks possess losses that differ from image captioning (or other downstream tasks), and thus IC models require further fine-tuning. Hence, recent work has focused on unifying generative vision-language models through text generation \cite{cho2021unifying,wang2022all, Wang2022b}, in order to optimize knowledge transfer from train to test. Lastly, novel methods have focused on optimally leveraging language caption supervision during pre-training, as small datasets with large caption variability can lead to detrimental effects \cite{Santurkar2023capt}.

\paragraph*{Symbolic Knowledge} Vision-language (VL) tasks can also be improved by incorporating symbolic knowledge into the VL models. For instance, providing a knowledge base, instantiated as subject-relation-object triplets associated with the images, both improve performance in vision-question answering (VQA) tasks, on top of allowing to explain the VQA model's predictions \cite{Riquelme2020}. In the same vein, adding high-level (semantic) attributes as inputs to IC models can increase captioning benchmarks \cite{you2016image,Yao2017}. Alternative efforts have shown that using object tags to facilitate the semantic image-text alignment during pre-training, and improves benchmark metrics in downstream fine-tuned image captioning tasks \cite{Li2020a}. Moreover, aligning directional semantic and spatial relationships between text and image (i.e., relation-level alignment) improves compositional reasoning \cite{Pandey2022}. 
Finally, symbolic knowledge and reasoning capability aim to enhance textual model's robustness when faced with out-of-distribution examples, thereby enabling them to engage in more human-like reasoning \cite{Collins2022}.



\paragraph*{Bias/Bug detection, and Evaluation} TIDA enhances the likelihood of simultaneously observing distinct attributes in an image within the augmented dataset. Thereby, our work relates to studies that focus on improving the predictive abilities of models in domains that suffer from bias-induced incorrect predictions. In line with this idea, the \textit{Equalizer} model is constrained to attend to the person attribute in images, increasing the IC abilities to detect the gender in the image \cite{Hendricks2018}. Interestingly, other attributes such as numeracy (e.g., counting) naturally emerge in standard embeddings \cite{Wallace2019numbers}, and may thus be less prone to biased predictions. Alternative debiasing methods focus on "decoupling" biased directions within text embeddings \cite{Chuang2023}. 

Other approaches focus on discovering the specific images where IC models fail (i.e., bugs). An instance of such a method uses a sequential pipeline that generates images from specific captions, classifies the object in the image, creates captions from the incorrectly classified images, generates captions of these images, and finally regenerates novel images based on the previously generated caption via a text-to-image generative process. These last images can be used to assess the robustness of vision models, as well as improve their performance \cite{Wiles2022}.

Moreover, while image captioning is usually scored on automatic metrics like SPICE \cite{anderson2016spice} or CIDEr \cite{vedantam2015cider}, it has been suggested that metrics evaluating both precision \textit{and} recall leading to better correlations with human judgments \cite{Kasai2022}. 
Finally, \cite{Hu2023} propose a method to compare image captioning models correlated with human judgment by leveraging LLM \cite{OpenAI2023}. 

\paragraph*{Data augmentation and Image generation} Data augmentation has been shown to improve performance both in vision \cite{ho2019population,cubuk2020randaugment} and language \cite{sennrich2015improving,Karimi2021a,andreas2019good,wei2019eda} tasks. Typically, data augmentation techniques involve procedures such as geometric transformations, color space augmentations, kernel filters, or
mixing images (see \cite{shorten2019survey} for review). To further improve these augmentation methods, a multi-task view of augmentation proposes to incorporate both original data and augmented examples during the training procedure \cite{Wei2021b}. This proposal has the benefit to relax the assumption that augmented examples cannot be too dissimilar from the original data. In the same vein, \textit{Neurocounterfactuals} is a method that allows augmenting data via large counterfactual perturbations that still bear resemblance to the original data but can nonetheless provide richer data augmentation \cite{Howard2022}. More recent studies have investigated data augmentation methods in multimodal settings such as VL tasks. For instance, LeMDA is a method that learns an augmentation network alongside a task-dedicated network \cite{liu2022learning}. This method augments the latent representation of the network and thus preserves the semantic structure in each modality. 

Moreover, not restricting data augmentation to the specificity of inputs can have detrimental effects, as augmented examples may possibly be associated to another label (e.g., a color change from green to red rock may induce a label change from emerald to ruby). To avoid this pitfall, instance-specific augmentation (\textit{InstaAug}) learns to apply invariances to specific parts of the input space \cite{Miao2022}. Similar work suggests estimating invariances by learning a distribution over augmentations, and jointly optimizing both the network \textit{and} augmentation distribution parameters \cite{Benton2020}. 

Other methods belong to a class of automated data augmentation algorithms. These algorithms can for example use reinforcement learning (RL) to optimize a data augmentation policy (e.g., \cite{cubuk2019autoaugment}). Furthermore, differentiable data augmentation proposes a method that relaxes the discrete state search assumption of RL, and allows for a more efficient data augmentation by implementing an end-to-end differentiable search procedure \cite{hataya2020faster}. Notably, other methods such as \textit{AdaAug} extend previous research by focusing not only on instance-depend data augmentation but also on class-dependent ones through the implementation of adaptive augmentation policies \cite{Cheung2022}.

Our method differentiates from policy-based methods for data augmentation but remains both automated, class-dependent, and targeted (i.e., we can focus on specific attributes such as gender, counting, or color). In particular, we leverage the impressive natural language-driven image synthesis abilities of text-to-image generative models \cite{yu2022scaling,saharia2022photorealistic,ramesh2022hierarchical} (see methods). In particular, we focus on their image editing or inpainting ability, which is a difficult challenge for these models given that only part of the image has to be changed while the rest has to be maintained. To solve this issue, traditional methods make use of explicit masks to circumscribe the inpainting region \cite{nichol2021glide,avrahami2022blended}. However, masking methods are both time-consuming and do not leverage structural information in the image. To circumvent this issue, recent work proposes the use of a prompt-to-prompt procedure in combination with a cross-attentional control mechanism that allows to edit of specific objects in the image while taking into account the contextual information \cite{Hertz2022}. Another method proposes to use of null-text inversion to achieve maskless image editing \cite{Mokady2022}. 

Interestingly, these state-of-the-art inpainting models open up the possibility to implement novel data augmentation methods. For instance, a recent paper showed that fine-tuning large-scale image-to-text generative models allows producing high-quality synthetic data that can improve ImageNet benchmark scores \cite{Azizi2023}. TIDA extends this idea in VL models, in order to improve specific target skills of these models within the framework of image captioning tasks. 


\section{Method and Experiments}

We propose a two-step method that allows retrieving certain images using their captions, regarding a specific concept that we call \textit{skill}. These skills refer to human- and animal-like basic abilities, such as gender categorization, counting, or recognizing colors. 
We first use a text mining method to detect whether or not a caption contains specific words that are related to the skill (Subsection \ref{subsec:skill}). 
Second, we generate variants of the original skill-related captions and create new images with these new captions in order to augment the dataset for each type of skill (Subsection \ref{subsec:tda}). An overview of the method is shown in Figure \ref{fig:tida}. 

\subsection{Skill-related retrieval} \label{subsec:skill}

We assume a list of $S$ skills $\{\mathcal{S}_i, i=1...S\}$, a training dataset of captions and images $\mathcal{D}^{\text{train}}=\{(\textbf{C}_k, I_k), k=1..k_{\text{train}}\}$, $\textbf{C}_k$ being a set of ground truth captions. 

For each skill $\mathcal{S}_i$ we create a binary classifier $f_{\mathcal{S}_i}$ that detects whether or not the skill $\mathcal{S}_i$ is present in a pair of image and associated captions. By applying this function to a dataset $\mathcal{D}$, it is possible to create a subpart of this dataset ${\mathcal{D}}_{\mathcal{S}_i}$ containing samples related to the aforementioned skill. 
By using this method and for each skill $\mathcal{S}_i$, we retrieve a subpart of the train ${\mathcal{D}^{\text{train}}}$  dataset that we call ${\mathcal{D}^{\text{train}}}_{\mathcal{S}_i}$ and a subpart of the test ${\mathcal{D}^{\text{test}}}$  dataset that we call ${\mathcal{D}^{\text{test}}}_{\mathcal{S}_i}$. The former will be used for data-augmentation and the latter will be used for the evaluation of the different models.

\begin{table*}
\centering
\resizebox{1.0\textwidth}{!}{
	\begin{tabular}{c|c|c|c|c|c||c|c|c|c||c|c|c}
         & \textbf{\#DA} & \multicolumn{4}{c||}{\textbf{BLEU@1-4}} & \multicolumn{4}{c||}{\textbf{RefCLIPScore}} & \multicolumn{3}{c}{\textbf{Spice}} \\
		\textbf{Test} & & \multirow{2}*{${\mathcal{D}^{test}}_{clr}$} & \multirow{2}*{${\mathcal{D}^{test}}_{ctg}$} & \multirow{2}*{${\mathcal{D}^{test}}_{gdr}$} & \multirow{2}*{${\mathcal{D}^{test}}$} & \multirow{2}*{${\mathcal{D}^{test}}_{clr}$} & \multirow{2}*{${\mathcal{D}^{test}}_{ctg}$} & \multirow{2}*{${\mathcal{D}^{test}}_{gdr}$} & \multirow{2}*{${\mathcal{D}^{test}}$}  & \multirow{2}*{F1$_{clr}$} & \multirow{2}*{F1$_{ctg}$} & \multirow{2}*{F1$_{all}$} \\ 
		\textbf{Train}	&  &   &     &     & &     &     &     &   &     &     &   \\ \hline \hline
			$\mathcal{D}^{train}$ (Vanilla)	& 0 & {51.8} & 44.0 & 49.9 & 49.7 &  79.9 & 79.3 & 79.8 & 80.3 & 24.1 & 19.7 & 20.7
 \\ \hline 
 	${\mathcal{D}^{train}}_{SD-rnd}$		& 60k &  51.3 & 44.1 & 49.2 & 49.6 & 80.0 & 79.5 & 79.7 & 80.2 & \textbf{24.7} & \textbf{25.2} & 20.6 \\ \hline 
	${\mathcal{D}^{train}}_{SD-clr}$	& 20k &   \textit{51.7} & 44.0 & \textit{49.3} & 49.5 &  79.8 & 79.4 & 79.6 & 80.1 &   24.3 & 19.8 & 20.2 \\ 
	${\mathcal{D}^{train}}_{SD-ctg}$	& 20k &   \textit{51.7} & \textit{44.4} & 49.2 & 49.7 &   79.9 & \textit{79.5} & 79.7 & 80.2 &  23.4 & 22.0 & 20.4
    \\ 
	${\mathcal{D}^{train}}_{SD-gdr}$	& 20k &    51.2 & 43.4 & 48.5 & 48.8 &   \textit{80.0} & 79.2 & \textit{79.9} & 80.3 &   24.5 & 24.4 & 20.6\\ \hline 
	${\mathcal{D}^{train}}_{SD-all}$		& 60k &    {51.8} & {44.9} & \textbf{50.1} & \textbf{50.5} &    {80.1} & \textbf{79.7} & {80.1} & \textbf{80.5} & \textbf{24.7} & 23.6 & \textbf{21.0} \\ \hline 
	\end{tabular}
 }
 \caption{Average of the BLEU@1-4 scores of the different TIDA-enhanced models on the different test sets. The TIDA models depicted used different image generation strategies: \textit{SD} uses Stable Diffusion and \textit{AAE} Attend-and-Excite. 
 The first line contains the performance of the model trained with the Vanilla train set. Then, the first to third line of each TIDA model contain the results of the model trained with data-augmentation on the color, counting, and gender skills, respectively. And, the last line of each, depicts the results of the model trained with all three types of data-augmentation. The scores in bold are the best scores on each test set, while the scores in italic are the best scores of each of the models trained with (skill-related) data-augmentation.} \label{tab:results}
\end{table*}

\subsection{Targeted Data Augmentation} \label{subsec:tda}

In order to improve the performances of the model with regard to several skills, we augment the dataset with sets of new examples. Those examples are created so that they depict new situations that are not necessarily in the training set, but should help the model generalize. 
For this purpose, we create a set of text generators functions 
$\{\mathcal{G}_{t, \mathcal{S}_i}, i=1 ... S\}$ 
taking as input a text caption containing a skill $\mathcal{S}_i$ and generating a slightly different version of this caption. 
The generator function perturbs the caption's text in such a way that it remains 
related to the skill. For example, it would inverse the gender of one of the words in the sentence: The caption "a man is playing basketball" would be changed (or perturbed) to "a woman is playing basketball". 
Mathematically, for any caption $c_{kl}$\footnote{caption $l$ of the image $k$} containing the skill ${\mathcal{S}_i}$, we create another caption $c_{kli}=\mathcal{G}_{t, \mathcal{S}_i}(c_{kl})$. 

Finally, for every perturbated caption $c_{kli}$ we use a text-to-image generator $\mathcal{G}_{V}$ in order to create an image $I_{kli}$ associated with the novel caption. We obtain an artificial set of image-caption pairs, which gives with the original images, the dataset ${\mathcal{D}^{train}}_{\mathcal{G}_{V}-\mathcal{S}_i}$. 

Those augmented datasets ${\mathcal{D}^{train}}_{\mathcal{G}_{V}-\mathcal{S}_i}$ are used to train several image captioning models, which should focus more on the specific skill $\mathcal{S}_i$. Each of the models is then evaluated on the different test sets ${\mathcal{D}^{test}}_{\mathcal{S}_i}$ which contain the pairs of images and list of captions that are related to the skill $\mathcal{S}_i$. The pseudo-code is visible in Algorithm \ref{alg:tida}.

 
\begin{algorithm}
\caption{The TIDA method on train}\label{alg:tida}
\begin{algorithmic}
\Require Skills $\mathcal{S}_i$, Textual skill detectors $f_{\mathcal{S}_i}$, Text generators $\mathcal{G}_{t, \mathcal{S}_i}$, Image generator $\mathcal{G}_{V}$, Train set $\mathcal{D}^{\text{train}}=\{(c_{kl}, I_k)\}$

\For{$i \; \text{in} \; 1...S$}

\State ${\mathcal{D}^{\text{train}}}_{\mathcal{G}_{V}-\mathcal{S}_i} \gets {\mathcal{D}^{\text{train}}}$ \Comment{Initialize}

\State ${\mathcal{D}^{\text{train}}}_{\mathcal{S}_i} \gets f_{\mathcal{S}_i}(\mathcal{D}^{\text{train}})$ \Comment{IC pairs with skill \textit{i}}


    \For{$ (c'_{kl}, I'_k) \; \text{in} \; {\mathcal{D}^{\text{train}}}_{\mathcal{S}_i}$} 
    
    \State $c'_{kli} \gets \mathcal{G}_{t, \mathcal{S}_i}(c'_{kl})$ \Comment{Caption perturbation}
    
    \State $I'_{kli} \gets \mathcal{G}_{V}(c'_{kli})$ \Comment{Image generation}
    
    \State ${\mathcal{D}^{\text{train}}}_{\mathcal{G}_{V}-\mathcal{S}_i} \gets {\mathcal{D}^{\text{train}}}_{\mathcal{G}_{V}-\mathcal{S}_i}\cup\{(c'_{kli}, I'_{kli})\}$ \Comment{Adding the new pair}
    \EndFor





\EndFor

\end{algorithmic}
\end{algorithm}

\subsection{Dataset} 

For the image captioning task, we use the Flickr30K \cite{Young2014}, which is composed of 31K photographs of everyday activities, events, and scenes harvested from Flickr and 159K captions. Each image is described independently by five annotators who are not familiar with the specific entities and circumstances depicted in them. 
We follow Karpathy’s split\footnote{\url{cs.stanford.edu/people/karpathy/deepimagesent/caption datasets.zip}} \cite{Karpathy2017}, which gives 29.8k/1k/1k images for train/val/test. 

\subsection{Methodology}

\paragraph{Skill used}
We augment the data regarding three basic human skills: gender detection, counting capability and color recognition. We focus on these skills for consistency with previous work \cite{anderson2016spice}, and because they are considered as essential and acquired early in humans and present in animals \cite{wang2010gender,dacke2008evidence,davis1982counting}. 

\paragraph{Text generation}
For each skill, and for each of the captions that were retrieved as containing it, we changed the caption text by using an alternative attribute of the targeted skill. For this, we employed a list of defined words that were related to the targeted skills. Each of the skill-related words has a list of other words that can be used as a replacement. 
For gender, masculine words like "man" were replaced by their feminine counterparts like "woman". For color, we swapped the different colors altogether. For counting, we either added or subtracted 1 to the detected written number in the sentence ($\pm1$). See Appendix \ref{app:regex} for more details. 

\paragraph{Baseline}
We compared our method with a data-augmentation that consists of generating images from random captions of the dataset. In this way, we aim to show that the improvement in different performances do not only come from having a larger training set, but also to have a larger and more diverse training set. In the following, we call this augmented training set  ${\mathcal{D}^{train}}_{SD-rnd}$. 

\subsection{Implementation details}

\paragraph*{Text generator} We used simple regular expressions to find the different attributes of each skill. The replacement words were chosen randomly within the list of possible alternatives. More details are available in Appendix \ref{app:regex}.\footnote{All our code will be made available after publication.} 


\paragraph*{Image generator} 

We test a classical text-to-image generation technique with Stable Diffusion \cite{rombach2022high} 
and generated 20k images per skill. 
For Stable Diffusion, we used the version 1.5\footnote{\url{https://huggingface.co/runwayml/stable-diffusion-v1-5}} as described in \cite{rombach2022high}, leveraging the Diffusers library for its implementation \cite{von-platen-etal-2022-diffusers}. We used a 16-bit floating-point data type and a guidance scale set at 8, which constrained the extent to which textual prompts generated the resultant images. The resolution of the generated images was 128 x 128 pixels. The remainder of the parameters were set as default, as specified by the Diffusers library. 
%
In the Appendix \ref{app:other_generators}, we show experiments with more generators. 

\paragraph*{Image captioning} We used the BLIP model \cite{Li2022} because of its state-of-the-art performances on Image Captioning, with a publicly available code and pre-trained weights. 
We kept the same original hyper-parameters, adjusting only the batch size from 32 to 24 and using the ViT Base model as the image encoder, due to hardware limitations.
For the training, we also kept the AdamW \cite{loshchilov2018decoupled} original optimization algorithm with an initial learning rate of $10^{-5}$ that is decreased through the training based on a $cos(\cdot)$ function until it reaches $0$.
In order to compare models with different amounts of available data, we used early stopping with a patience of $5$. 

\paragraph*{Metrics}
We used the classical BLEU metric \cite{Papineni2002} to evaluate the performances of the models. 
Moreover, we used another metric that relies on learned representations. We computed RefCLIPScore \cite{Hessel2021} which is based on the similarity between the embedding of the caption and the embedding of the image coming from CLIP \cite{Radford2021}. This metric was shown to have a better correlation with human judgments than other classical metrics \cite{Kasai2022}. 



\section{Results and Analysis} 

\subsection{Results}

\begin{table*}
    \resizebox{1.0\textwidth}{!}{
    \centering
	\begin{tabular}{c|ccccc|ccccc|ccccc}
		\textbf{Skill} & \multicolumn{5}{c|}{\textbf{Color}} & \multicolumn{5}{c|}{\textbf{Counting}} & \multicolumn{5}{c}{\textbf{Gender}}   \\ 
		\textbf{Train}	&  P+   &  R+   &   P-   &  R-   & F1 &  P+   &  R+   &  P-   &  R-   & F1 &  P+   &  R+ &  P-   &  R-   & F1  \\ \hline \hline
			$\mathcal{D}^{train}$	&   64.4 & 89.8 & 80.5 & 45.8 & 66.7 & 73.6 & 97.9 & 91.7 & 39.1 & 69.4 & 46.5 & 88.8 & 97.2 & 79.0 & \textbf{74.1}
 \\ \hline 
     ${\mathcal{D}^{train}}_{SD-rnd}$ & 64.8 & 88.1 & 78.6 & 47.7 & 67.0 & 77.2 & 97.5 & 92.0 & 50.0 & \textbf{75.5} & 45.4 & 89.4 & 97.3 & 78.0 & 73.4 \\ \hline 
	${\mathcal{D}^{train}}_{SD-clr}$	& 66.0 & 86.8 & 78.0 & 51.3 & \textbf{68.4} & 73.4 & 98.4 & 93.3 & 38.3 & 69.2 & 43.8 & 91.8 & 97.8 & 75.9 & 72.4 \\ 
	${\mathcal{D}^{train}}_{SD-ctg}$	&  65.5 & 88.5 & 79.7 & 49.2 & 68.1 & 74.4 & 98.1 & 92.7 & 41.5 & 71.0 & 44.8 & 91.8 & 97.9 & 76.9 & 73.2
    \\ 
	${\mathcal{D}^{train}}_{SD-gdr}$	&  64.1 & 88.5 & 78.5 & 45.8 & 66.1 & 75.3 & 96.8 & 89.2 & 45.1 & 72.3 & 43.9 & 90.6 & 97.5 & 76.3 & 72.4\\ \hline 
	${\mathcal{D}^{train}}_{SD-all}$		&    65.7 & 90.8 & 82.8 & 48.3 & \textbf{68.6} & 75.8 & 97.8 & 92.3 & 45.9 & \textbf{73.4} & 46.0 & 92.4 & 98.0 & 77.8 & \textbf{74.1} \\ 
	\end{tabular}
 }
 \caption{Precision, Recall and F1-score regarding the use of skill-related words in the captions generated by the BLIP models trained using different TIDA techniques on the different test sets. The two best F1 scores are highlighted in bold.
 } \label{tab:fscores}
\end{table*}

The results of the models trained with different skill-based data-augmentation on different test sets are shown in Table \ref{tab:results}. 
%
%
We can see 
that the overall best scores on each test set are obtained with the model using the three types of data-augmentation techniques, either using BLEU (from 49.7 to 50.5) or RefCLIPScore (from 80.3 to 80.5).

We also provide the F1-scores computed with Spice, and especially the ones related to counting and color because we aim to quantify the performances of the models on those skills. The data-augmentation helps to augment both of the metrics individually, more than the overall one. 

\subsection{Analysis}

We analyze the results in three different ways: (i) by using classical natural language generation metrics for image captioning, (ii) by assessing the use of skill words regarding the captions and quantifying the right use of the skill-related terms, 
(iii) by probing the representation of the image on a skill detection task for a finer comprehension of the image encoder and text decoder behavior. 

\paragraph{Classical metrics}
By analyzing the classical metrics we can make several observations. Contrary to what we would have expected, the skill-related TIDA are not necessarily leading to the best scores in their respective test sets. Note however that the metrics are not homogeneous. 
The counting-related TIDA obtains the best results on the counting test set for BLEU and RefCLIPScore, but Spice F1-counting is better with gender. Interestingly, counting (compared with color and gender) leads to the worst metrics with BLEU but the best one when focusing on the RefCLIPScore and Spice metrics. 
More details and metrics are available in Appendix \ref{app:metrics}.

\paragraph{Skill-related words}

In order to analyze the results of the model by going beyond the classical opaque metrics like BLEU and RefCLIPScore, we used a similar method to spice \cite{anderson2016spice} that allows to investigate specific semantic words. 
TIDA relies on using certain variations of words, hence we are evaluating the propensity of the model to use those words in the right context. 
If a word associated with a skill is present in the ground truth or in the generated caption, it allows us to quantify the results of the model as false/true positive/negative.  
Specifically, when the model is using a word associated with a skill in the generated caption, and this skill is indeed associated with the image-caption ground truth, we count this as a true positive. If the model does not use any word associated with a skill and the skill is not present in the ground truth, we count this as a true negative. The other combinations are regarded as false positives or negatives. 
The precision, recall, and F1 for color, counting, and gender TIDAs are available in Table \ref{tab:fscores}. 

For the color TIDA, the precision and recall are both increasing for the positive and negative cases. 
This means that the model is using more often color words when the caption should contain one and less when it should not. 
For the counting TIDA, the recall of the negative class is augmenting from 39.1 to 45.9, which means that the model uses fewer counting-related words when it should not. At the same time the precision for the positive class augments which means the use of counting-related words is more pertinent. 
For the gender TIDA, the model is using more gender words (recall positive going from 88.8 to 92.4) while being a bit less precise (recall negative decreasing from 79.0 to 77.8). 
Overall, we observe that the color TIDA gives better results for color, but surprisingly the counting TIDA is better for gender and the gender TIDA is better for counting.   

\begin{table}
    \centering
    \hspace*{-.4cm}
    \begin{tabular}{c|l|l|l}
     \textbf{Skill} & \multicolumn{1}{c|}{\textbf{Color}} & \multicolumn{1}{c|}{\textbf{Counting}} & \multicolumn{1}{c}{\textbf{Gender}} \\ \hline \hline
    $\mathcal{D}^{train}$            &   72.0 &     88.2 &     84.1  \\ \hline
    ${\mathcal{D}^{train}}_{SD-rnd}$ &   73.0 &     88.3 &     84.3  \\ \hline
    ${\mathcal{D}^{train}}_{SD-clr}$ &   72.9 &     88.6 &     84.7  \\
    ${\mathcal{D}^{train}}_{SD-ctg}$ &   71.6 &     88.7 &     84.1  \\
    ${\mathcal{D}^{train}}_{SD-gdr}$ &   71.7 &     89.0 &     84.0  \\ \hline
    ${\mathcal{D}^{train}}_{SD-all}$ &   71.8 &     87.7 &     84.3  \\ \hline

    

    \end{tabular}
    \caption{F1-score for skill probing using the models learned with different targeted data-augmentations}
    \label{tab:skill-probing}
\end{table}


%



\paragraph{Probing with visual representations}
We tried to analyze how TIDA influences the model not only using the raw results of the text decoder but also using the representation of the image encoder.
For this purpose, we proposed to probe the image representations to predict whether or not the image is associated with a specific skill.

As we previously did, we used the text-mining method to label whether or not a sample is associated with one of the three skills.
We then trained a linear multi-layer perceptron on the representations produced by the image encoder and these labels.
As is usual with transformer-based models, we used the class embedding coming from the image encoder as the image representation embedding.
We use binary cross entropy loss and SGD to train the probe and perform early stopping and a grid search on each model to find the best model hidden size and learning rate.
The results with the five TIDA models are shown in Table \ref{tab:skill-probing}.

Looking at the F1-score, it seems that none of the TIDAs bring any significant change regarding the skill-related information in the image encoding.
However, the models are improving in terms of general Image Captioning performances (Table \ref{tab:results}), and we saw previously that they are using more frequently targeted words when they should use them (Table \ref{tab:fscores}).
We can conclude that TIDA-related improvements are caused by changes in the text decoder rather than the image decoder.


\section{Conclusion and Future Work}

This paper assesses the effectiveness of generative data augmentation with current diffusion models for improving specific skills of image captioning models. We use the Flickr30k image captioning dataset and ran experiments with BLIP, a recent vision-language state-of-the-art model. We show that TIDA, our targeted image data-augmentation techniques allows for gains regarding classical metrics that are recognized by the community like BLEU or RefCLIPScore. 
On top of that, we also propose a fine-grained analysis to analyze the results of the model by going beyond the classical opaque metrics by investigating the occurrences of specific semantic words related to the target skills. We found out that TIDA helps the image captioning model to use those words more efficiently. 
Finally, we investigate the visual part, we probe the representations from the visual encoder and reveal  that they do not contain more information on the skill, meaning the improvement relies on the textual decoder. 

Our results open several avenues for further research. For instance, it remains unclear why we observe the boost in results on a specific skill when using data-augmentation on another skill. 
It would also be useful to investigate more in details the reasons of the improvement of performances the text decoder or the visual encoder, or to use a more precised metric powered by a LLM like \cite{Hu2023}. 

It would also be useful to investigate more in details the reasons of the gain of performances of the text decoder or the visual encoder, or to using complex interpretable metrics from LLM like the Text-to-Image Faithfulness Evaluation with Question Answering \cite{Hu2023}. 
It would be to see improvements with text-to-image models known to be better at generating images related to color, counting, like Attend-and-Excite \cite{Chefer2023} with newer versions of stable diffusion. 
Finally, we would like to extend our method to Visual Question Answering. Using symbolic knowledge to extract the objects of the image-caption and the relation as implemented in \cite{Riquelme2020}, we can adapt the model to new situations and help to de-bias a VQA model. 
Finally, given the recent results of \cite{Azizi2023}, we should run a random data-augmentation on the train set and see whether this procedure may help to improve the results compared with TIDA. 

\section{Limitations}

The focus of this work has been on abstract skills shown to be learned by humans at an early age, 
but it is not clear which skills are the most important to image captioning in particular or another particular task in general.  
And it is an empirical study to determine which skills result in the most improvement in a task.
Making it not straightforward to add new skills, requiring thoughtfulness and empirical validation.

In terms of computational cost, TIDA's necessity to generate a number of new examples comparable to the original dataset size using costly neural image generation models signifies it is a challenge to apply to larger datasets and that the technique doesn't scale well to dataset size. 
And although each generated example can be leveraged many times, the process is heavily limited by the computation capabilities.

\section*{Acknowledgments}
This work was funded by National Center for Artificial Intelligence CENIA FB210017, Basal ANID.

\bibliography{emnlp2023-latex/ijcai23,emnlp2023-latex/val,emnlp2023-latex/JRC}
\bibliographystyle{acl_natbib}


\appendix

\section{List of skill-related words} 
\label{app:regex}

\paragraph{Color} We used seven colors: blue, brown, green, grey, orange, pink, purple, red, and yellow. We inverted them randomly.

\paragraph{Counting} We used all the numbers from one to six. All the captions only contained written numbers.

\paragraph{Gender} For male, we used the words boy, boys, man, men, guy, and guys. They were changed with the words girl, girls, woman, and women. 

\section{Other Image Generators} 
\label{app:other_generators}

 We generate the images with different techniques. In-Painting mode, in order to change the images the less possible, and another image generator algorithm called Attend-and-Excite \cite{Chefer2023}, in order to stress specific tokens of the sentence used to generated, related to the attribute we want to enhance. Results are in Tables \ref{tab:results_IP} and \ref{tab:fscores_IP}.

\begin{table*}[!h]
\centering
\resizebox{1.0\textwidth}{!}{
	\begin{tabular}{c|c|c|c|c||c|c|c|c||c|c|c}
         & \multicolumn{4}{c||}{\textbf{BLEU@1-4}} & \multicolumn{4}{c||}{\textbf{RefCLIPScore}} & \multicolumn{3}{c}{\textbf{Spice}} \\
		\textbf{Test} & \multirow{2}*{${\mathcal{D}^{test}}_{clr}$} & \multirow{2}*{${\mathcal{D}^{test}}_{ctg}$} & \multirow{2}*{${\mathcal{D}^{test}}_{gdr}$} & \multirow{2}*{${\mathcal{D}^{test}}$} & \multirow{2}*{${\mathcal{D}^{test}}_{clr}$} & \multirow{2}*{${\mathcal{D}^{test}}_{ctg}$} & \multirow{2}*{${\mathcal{D}^{test}}_{gdr}$} & \multirow{2}*{${\mathcal{D}^{test}}$}  & \multirow{2}*{F1$_{clr}$} & \multirow{2}*{F1$_{ctg}$} & \multirow{2}*{F1$_{all}$} \\ 
		\textbf{Train}	&     &     &     & &     &     &     &   &     &     &   \\ \hline \hline
		$\mathcal{D}^{train}$	&  {51.8} & 44.0 & 49.9 & 49.7 &  79.9 & 79.3 & 79.8 & 80.3 & 24.1 & 19.7 & 20.7 \\ \hline 
    ${\mathcal{D}^{train}}_{INP-clr}$	&  51.4 & 44.8 & \textit{49.8} & \textit{50.1} & 79.8 & 79.1 & 79.6 & 80.1 & 23.1 & 20.1 & 20.4  \\
    ${\mathcal{D}^{train}}_{INP-ctg}$	&  \textit{\textbf{52.2}} & \textit{\textbf{45.1}} & 49.3 & 49.8 & 80.2 & 79.3 & 79.7 & 80.2 & \textit{\textbf{25.2}} & 21.3 & 20.6  \\
    ${\mathcal{D}^{train}}_{INP-gdr}$	&  50.9 & 42.8 & 48.3 & 48.7 & \textit{\textbf{80.3}} & \textit{79.6} & \textit{\textbf{80.2}} & \textit{\textbf{80.5}} & 23.1 & \textit{22.4} & \textit{20.7}  \\ \hline
    ${\mathcal{D}^{train}}_{INP-all}$	&  51.3 & 44.0 & 49.2 & 49.5 & 79.7 & 79.0 & 79.6 & 80.1 & 23.9 & 21.3 & 20.4  \\ \hline \hline

    ${\mathcal{D}^{train}}_{AAE-clr}$ & 51.7 & 42.8 & 48.7 & 49.1 & 80.0 & 79.0 & 79.7 & 80.2 & 22.6 & 20.8 & 20.5  \\
    ${\mathcal{D}^{train}}_{AAE-ctg}$  & 52.1 & 44.6 & 49.7 & 49.9 & 79.8 & 79.2 & 79.7 & 80.2 & 24.6 & 20.3 & 20.5  \\
    ${\mathcal{D}^{train}}_{AAE-gdr}$  & 51.4 & 43.5 & 49.3 & 49.4 & 80.1 & 79.4 & 80.1 & \textbf{80.5} & 23.7 & 19.2 & 20.5  \\ \hline 
    ${\mathcal{D}^{train}}_{AAE-all}$ & 51.1 & 43.4 & 48.8 & 49.1 & 79.9 & 79.5 & 80.1 & 80.4 & 22.9 & 20.7 & \textbf{21.0 } \\
    
	\end{tabular}
 }
 \caption{Average of the BLEU@1-4 scores of the different TIDA-enhanced models on the different test sets. The TIDA models depicted used different image generation strategies: \textit{SD} uses Stable Diffusion, \textit{AAE} Attend-and-Excite, and \textit{INP} Inpaiting. The first line contains the performance of the model trained with the Vanilla train set. Then, the first to third line of each TIDA model contain the results of the model trained with data-augmentation on the color, counting, and gender skills, respectively. And, the last line of each, depicts the results of the model trained with all three types of data-augmentation. The scores in bold are the best scores on each test set, while the scores in italic are the best scores of each of the models trained with (skill-related) data-augmentation.  } \label{tab:results_IP}
\end{table*}

\begin{table*}
    \resizebox{1.0\textwidth}{!}{
    \centering
	\begin{tabular}{c|ccccc|ccccc|ccccc}
		\textbf{Skill} & \multicolumn{5}{c|}{\textbf{Color}} & \multicolumn{5}{c|}{\textbf{Counting}} & \multicolumn{5}{c}{\textbf{Gender}}   \\ 
		\textbf{Train}	&  P+   &  R+   &   P-   &  R-   & F1 &  P+   &  R+   &  P-   &  R-   & F1 &  P+   &  R+ &  P-   &  R-   & F1  \\ \hline \hline
			$\mathcal{D}^{train}$	&   64.4 & 89.8 & 80.5 & 45.8 & 66.7 & 73.6 & 97.9 & 91.7 & 39.1 & 69.4 & 46.5 & 88.8 & 97.2 & 79.0 & 74.1
 \\ \hline 
 
${\mathcal{D}^{train}}_{INP-clr}$ & 63.6 & 91.2 & 81.7 & 42.9 & 65.6 & 73.3 & 98.4 & 93.3 & 38.0 & 69.0 & 45.1 & 89.4 & 97.3 & 77.7 & 73.2 \\
${\mathcal{D}^{train}}_{INP-ctg}$ & 64.7 & 87.9 & 78.4 & 47.7 & 66.9 & 74.5 & 96.8 & 88.6 & 42.6 & 70.9 & 42.6 & 91.8 & 97.8 & 74.7 & 71.5 \\
${\mathcal{D}^{train}}_{INP-gdr}$ & 63.1 & 88.7 & 77.8 & 43.3 & 64.7 & 74.4 & 96.8 & 88.6 & 42.3 & 70.7 & 44.7 & 90.0 & 97.4 & 77.2 & 73.0 \\ \hline
${\mathcal{D}^{train}}_{INP-all}$ & 64.5 & 88.9 & 79.4 & 46.7 & 66.8 & 74.3 & 97.8 & 91.6 & 41.5 & 70.8 & 45.8 & 92.9 & 98.2 & 77.5 & 74.0 \\ \hline \hline

    ${\mathcal{D}^{train}}_{AAE-clr}$ & 62.8 & 90.4 & 79.9 & 41.6 & 64.5 & 74.3 & 97.5 & 90.5 & 41.5 & 70.6 & 47.4 & 91.2 & 97.8 & 79.3 & 75.0 \\
    ${\mathcal{D}^{train}}_{AAE-ctg}$ & 64.0 & 88.7 & 78.6 & 45.4 & 65.9 & 74.0 & 98.4 & 93.6 & 40.2 & 70.4 & 47.3 & 91.2 & 97.8 & 79.2 & 74.9 \\
    ${\mathcal{D}^{train}}_{AAE-gdr}$ & 63.9 & 90.0 & 80.3 & 44.4 & 65.9 & 74.3 & 97.8 & 91.6 & 41.5 & 70.8 & 42.9 & 90.0 & 97.4 & 75.4 & 71.5 \\ \hline
    ${\mathcal{D}^{train}}_{AAE-all}$ & 64.4 & 90.6 & 81.5 & 45.2 & 66.7 & 75.4 & 97.3 & 90.7 & 45.1 & 72.6 & 48.6 & 90.6 & 97.7 & 80.4 & 75.7 \\ 
 
	\end{tabular}
 }
 \caption{Precision, Recall and F1-score regarding the use of skill-related words in the captions generated by the BLIP models trained using different TIDA techniques on the different test sets} \label{tab:fscores_IP}
\end{table*}

\subsection{In Painting Model}

We ran more experiments with another configuration for image generation that we call Inpainting (INP). It consists of changing only a subpart of the initial image in order to perturbate it. 
For this configuration, we first segmented the desired object in the scene by using a pretrained ClipSeg model \cite{lueddecke:22}, by prompting the nominal group of the skill-related word. The segmentation mask was obtained by setting an element-wise threshold of $0.1$ in the final output of the model, after applying sigmoid and a min-max normalization. The mask was then dilated using a square kernel of $10$ x $10$ pixels. The original image was finally inpainted using the pretrained model of \citep{rombach:22}. 

\subsection{Attend-and-Excite}

We tried to change the classical stable diffusion by another version called Attend-and-Excite (\textit{AAE}; \citealp{Chefer2023}), which enhance the classical stable diffusion model to make it better at generating specific attribute. 

We used the model described in \cite{Chefer2023}, using as backbone the version 1.5\footnote{\url{https://huggingface.co/runwayml/stable-diffusion-v1-5}} of stable diffusion, with the official implementation of the authors which is also built on top of the Diffusers library. The default parameters were used as default, expect regarding the number maximum of refinement steps, which has been downgraded from 20 to 5. 


\section{Other metrics} \label{app:metrics}

Results using other metrics are shown in the section. Table \ref{tab:spice} and Table \ref{tab:cider} contain respectively the results with Spice and Cider. 

\begin{table}
    \resizebox{1.0\linewidth}{!}{
	\begin{tabular}{c|c|c|c|c}
		\textbf{Test} & ${\mathcal{D}^{test}}_{clr}$ & ${\mathcal{D}^{test}}_{ctg}$ & ${\mathcal{D}^{test}}_{gdr}$ & ${\mathcal{D}^{test}}$ \\ 
		\textbf{Train}	&     &     &     &  \\ \hline \hline
			$\mathcal{D}^{train}$	&  21.3 & 18.5 & 20.3 & 20.7 \\ \hline 
			${\mathcal{D}^{train}}_{SD-rnd}$	&  21.4 & 18.2 & 20.1 & 20.6 \\ \hline 
	${\mathcal{D}^{train}}_{SD-clr}$	&   20.9 & 17.9 & 19.7 & 20.2 \\ 
	${\mathcal{D}^{train}}_{SD-ctg}$	&  21.0 & 18.2 & 20.0 & 20.4\\ 
	${\mathcal{D}^{train}}_{SD-gdr}$	&   20.8 & 18.8 & 19.9 & 20.6 \\ \hline 
	${\mathcal{D}^{train}}_{SD-all}$		&   21.0 & 19.3 & 20.3 & 21.0 \\ \hline

${\mathcal{D}^{train}}_{AAE-clr}$ & 20.8 & 18.0 & 19.8 & 20.5 \\
${\mathcal{D}^{train}}_{AAE-ctg}$ & 21.1 & 18.6 & 20.0 & 20.5 \\
${\mathcal{D}^{train}}_{AAE-gdr}$ & 21.0 & 18.3 & 19.9 & 20.5 \\ \hline
${\mathcal{D}^{train}}_{AAE-all}$ & 21.2 & 18.7 & 20.3 & 21.0 \\ \hline
${\mathcal{D}^{train}}_{INP-clr}$ & 20.7 & 18.4 & 19.9 & 20.4 \\
${\mathcal{D}^{train}}_{INP-ctg}$ & 21.6 & 18.8 & 20.2 & 20.6 \\
${\mathcal{D}^{train}}_{INP-gdr}$ & 21.1 & 18.9 & 20.1 & 20.7 \\ \hline
${\mathcal{D}^{train}}_{INP-all}$ & 20.9 & 18.4 & 19.9 & 20.4 \\
 
	\end{tabular}
 }
 \caption{Average of the Spice F1 scores of the different models on the different test sets 
 } \label{tab:spice}
\end{table}

\begin{table}
    \resizebox{1.0\linewidth}{!}{
	\begin{tabular}{c|c|c|c|c}
		\textbf{Test} & ${\mathcal{D}^{test}}_{clr}$ & ${\mathcal{D}^{test}}_{ctg}$ & ${\mathcal{D}^{test}}_{gdr}$ & ${\mathcal{D}^{test}}$ \\ 
		\textbf{Train}	&     &     &     &  \\ \hline \hline
			$\mathcal{D}^{train}$	&  102.5 & 81.1 & 95.3 & 99.6
 \\ \hline 
			${\mathcal{D}^{train}}_{SD-rnd}$	&  100.9 & 81.7 & 94.9 & 99.3
 \\ \hline 
	${\mathcal{D}^{train}}_{SD-clr}$	&   102.2 & 80.3 & 94.0 & 98.8 \\ 
	${\mathcal{D}^{train}}_{SD-ctg}$	&  102.2 & 82.3 & 93.9 & 99.0

    \\ 
	${\mathcal{D}^{train}}_{SD-gdr}$	&   100.1 & 81.9 & 92.7 & 98.0 \\ \hline 
	${\mathcal{D}^{train}}_{SD-all}$		&   101.0 & 81.4 & 95.7 & 101.5 \\ \hline


${\mathcal{D}^{train}}_{AAE-clr}$ & 102.2 & 77.8 & 92.7 & 98.0 \\
${\mathcal{D}^{train}}_{AAE-ctg}$ & 101.7 & 82.0 & 95.1 & 100.5 \\
${\mathcal{D}^{train}}_{AAE-gdr}$ & 99.5 & 78.1 & 93.6 & 98.0 \\ \hline
${\mathcal{D}^{train}}_{AAE-all}$ & 99.5 & 78.5 & 92.8 & 98.2 \\ \hline

${\mathcal{D}^{train}}_{INP-clr}$ & 100.8 & 82.9 & 95.3 & 100.5 \\
${\mathcal{D}^{train}}_{INP-ctg}$ & 104.5 & 83.7 & 94.7 & 99.8 \\
${\mathcal{D}^{train}}_{INP-gdr}$ & 101.7 & 80.6 & 94.1 & 99.0 \\ \hline
${\mathcal{D}^{train}}_{INP-all}$ & 100.7 & 82.3 & 94.5 & 99.4 \\
    
	\end{tabular}
 }
 \caption{Average of the Cider scores of the different models on the different test sets 
 } \label{tab:cider}
\end{table}

\section{Probing}

More results on the probing experiments are shown in Table \ref{tab:app-skill-probing}. 

\begin{table}
    \centering
    \hspace*{-.4cm}
    \resizebox{.55\textwidth}{!}{
    \begin{tabular}{c|lll|lll|lll}
     \textbf{Skill} & \multicolumn{3}{c|}{\textbf{Color}} & \multicolumn{3}{c|}{\textbf{Counting}} & \multicolumn{3}{c}{\textbf{Gender}} \\
    \textbf{Train}                      & P & R & F1         &       P & R & F1            & P & R &  F1 \\ \hline \hline
    $\mathcal{D}^{train}$               & 67.5 & 77.2 & 72.0 &     87.9 & 88.6 & 88.2      & 83.1 & 85.1 & 84.1  \\ \hline

    ${\mathcal{D}^{train}}_{SD-rnd}$ &      70.7 &   75.4 &     73.0 &     86.1 &   90.5 &     88.3 &      83.2 &   85.4 &      84.3 \\ \hline

    ${\mathcal{D}^{train}}_{SD-clr}$	& 69.1 & 77.2 & 72.9 &     86.0 & 91.4 & 88.6      & 83.3 & 86.2 & 84.7  \\
    ${\mathcal{D}^{train}}_{SD-ctg}$	& 66.3 & 77.8 & 71.6 &     85.1 & 92.6 & 88.7      & 82.6 & 85.7 & 84.1  \\
    ${\mathcal{D}^{train}}_{SD-gdr}$	& 67.8 & 76.1 & 71.7 &     85.5 & 92.7 & 89.0      & 83.9 & 84.2 & 84.0  \\ \hline
    ${\mathcal{D}^{train}}_{SD-all}$	& 60.1 & 89.1 & 71.8 &     86.8 & 88.6 & 87.7      & 83.3 & 85.3 & 84.3  \\ \hline

    ${\mathcal{D}^{train}}_{AAE-clr}$	& 68.5 & 75.9 & 72.0 &     86.7 & 89.2 & 88.0      & 84.1 & 86.5 & 85.3  \\
    ${\mathcal{D}^{train}}_{AAE-ctg}$	& 65.3 & 83.5 & 73.3 &     86.1 & 90.6 & 88.3      & 82.9 & 86.7 & 84.7  \\
    ${\mathcal{D}^{train}}_{AAE-gdr}$	& 71.8 & 73.7 & 72.7 &     85.2 & 91.9 & 88.4      & 84.0 & 86.7 & 85.3  \\ \hline
    ${\mathcal{D}^{train}}_{AAE-all}$	& 72.5 & 75.6 & 74.0 &     89.0 & 90.2 & 89.6      & 81.4 & 87.8 & 84.5  \\ \hline

    ${\mathcal{D}^{train}}_{INP-clr}$	& 63.7 &   80.5 & 71.1 &     84.3 &   91.0 & 87.5      & 84.6 &   83.4 & 84.0  \\
    ${\mathcal{D}^{train}}_{INP-ctg}$	& 67.6 &   79.1 & 72.9 &     88.1 &   89.1 & 88.6      & 83.9 &   84.8 & 84.3  \\
    ${\mathcal{D}^{train}}_{INP-gdr}$	& 66.0 &   81.0 & 72.7 &     88.6 &   89.5 & 89.0      & 82.2 &   85.9 & 84.0  \\ \hline
    ${\mathcal{D}^{train}}_{INP-all}$	& 66.4 &   79.4 & 72.3 &     87.4 &   91.1 & 89.2      & 85.9 &   83.7 & 84.8  \\ \hline

    \end{tabular}
    }
    \caption{Skill Probing}
    \label{tab:app-skill-probing}
\end{table}

\end{document}